\documentclass[runningheads]{llncs}
\usepackage[utf8]{inputenc}
\usepackage[english,british,USenglish]{babel}
\usepackage{float}
\usepackage{ifthen}
\usepackage{xspace}
\usepackage{wrapfig}
\usepackage{xstring}
\usepackage{csquotes}
\usepackage{footnote}
\usepackage{hyphenat}
\usepackage{multirow}
\usepackage{tabularx}
\usepackage{graphicx}
\usepackage{microtype}
\usepackage{marginnote}
\usepackage{subcaption}
\usepackage[all]{nowidow}
\usepackage{tablefootnote}
\usepackage[inline]{enumitem}
\usepackage[usenames,dvipsnames]{xcolor}
\usepackage{pgfplots}

\usepackage{varioref}
\usepackage{hyperref}
\usepackage{cleveref}
\hypersetup{
	bookmarksopen=true,
	bookmarks=true,
	hidelinks=true,
	colorlinks=true,
	linkcolor={green!50!black},
	citecolor={blue!50!black},
	urlcolor={blue!80!black}
}

\usepackage{bookmark}

\newcommand\citereq[1]{%
	\ifthenelse{\equal{#1}{}}{%
		\textcolor{red}{[Citation Required!]}%
	}{%
		\textcolor{RedOrange}{[Need More Citations!]}%
		\cite{#1}%
	}%
}

\newcommand{\formatmn}[1]{%
		\fontfamily{ptm}\selectfont%
		\bfseries\scriptsize#1%
}

\newcommand{\kword}[1]{%
	\marginnote{\formatmn{#1}}%
	\textbf{\nohyphens{\small#1}}%
	\xspace%
}

\newcommand\athome{RoboCup@Home\xspace}

\makeatletter
\renewcommand{\marginnote}[1]{}
\newcommand{\acronym}[3]{\newcommand{#1}[1][]{%
	\ifcsname @@#3\endcsname%
		\IfStrEqCase{##1}{%
			{n}{#3\marginnote{\formatmn{#3}}}%
			{f}{#2 (#3)}%
			{fn}{#2 (#3\marginnote{\formatmn{#3}})}%
			{x}{#2}%
			{xn}{#2\marginnote{\formatmn{#3}}}%
		}[#3]%
	\else%
		\IfSubStr{##1}{n}%
			{#2 (#3\marginnote{\formatmn{#3}})}%
			{#2 (#3)}%
		\expandafter\def\csname @@#3\endcsname{}%
	\fi\xspace%
}}
\makeatother

\acronym{\tc}{Technical Committee}{TC}
\acronym{\tdp}{Team Description Paper}{TDP}
\acronym{\tdps}{Team Description Papers}{TDPs}
\acronym{\hri}{Human-Robot Interaction}{HRI}
\acronym{\asr}{Automatic Speech Recognition}{ASR}
\acronym{\nlp}{Natural-Language Processing}{NLP}
\acronym{\nlu}{Natural-Language Understanding}{NLU}
\acronym{\ssl}{Sound Source Localization}{SSL}
\acronym{\spl}{Standard Platform Leagues}{SPL}
\acronym{\opl}{Open Platform League}{OPL}
\acronym{\dspl}{Domestic Standard Platform League}{DSPL}
\acronym{\sspl}{Social Standard Platform League}{SSPL}
\acronym{\gpsr}{\emph{General Purpose Service Robot}}{GPSR}
\acronym{\srad}{\emph{Speech Recognition and Audio Detection}}{SRAD}
\acronym{\egpsr}{\emph{Endurance General Purpose Service Robot}}{EGPSR}
\acronym{\eegpsr}{\emph{Enhanced Endurance General Purpose Service Robot}}{EEGPSR}

\title{%
From Commands to Goal-based Dialogs:\\%
A Roadmap to Achieve Natural Language Interaction in \athome%
}
\titlerunning{A Roadmap to Achieve Natural Language Interaction in \athome}
\author{%
	Mauricio Matamoros%
	\and%
	Karin Harbusch%
	\and%
	Dietrich Paulus%
}
\authorrunning{M. Matamoros et al.}
\institute{%
	Active Vision Group (AGAS), University of Koblenz-Landau.\\
	Universitätsstr. 1, 56070 Koblenz, Germany%
}

\begin{document}
\maketitle

%
%
%
%
%
%

\begin{abstract}

On the one hand, speech is a key aspect to people's communication.
On the other, it is widely acknowledged that language proficiency is related to intelligence.
Therefore, intelligent robots should be able to understand, at least, people's orders within their application domain.
These insights are not new in \athome, but we lack of a long-term plan to evaluate this approach.

In this paper we conduct a brief review of the achievements on automated speech recognition and natural language understanding in \athome.
Furthermore, we discuss main challenges to tackle in spoken human-robot interaction within the scope of this competition.
Finally, we contribute by presenting a pipelined road map to engender research in the area of natural language understanding applied to domestic service robotics.
\keywords{%
	Robotic competitions \and
	Natural Language Understanding \and
	Artificial intelligence and robotics%
	.%
}%
\end{abstract}

%
%
%
%
%

\section{Introduction}
\label{sec:introduction}

From its foundation, \athome has stressed the importance of natural interaction between humans and robots.
With the target perspective that intelligent robots can understand people's orders that fall within their application domain.
Thus, \hri has always been pursued in the competition.

However, a detailed evaluation of natural language interactions is not easy.
In favor of many other functions in early stages of the robot development, simple straightforward commands often served the purpose of interaction.
This, along with
\begin{enumerate*}[label=\alph*)]
	\item the noisy competition environments,
	\item the biased, non-native speaker operators;
	and
	\item the use of command generators to instruct robots
\end{enumerate*}
have impeded the definition of fine-grained evaluation measures \athome for natural language understanding in the domain of \athome.

In order to achieve this goal, we first analyzed the progression of the league in \asr and \nlu in the last nine years. Our proposal is based on claims made by the participating teams in their \tdps, relevant publications, rulebooks, multimedia material available on-line, and our cumulative experience as participants and referees in \athome since 2009.

In this paper we present a road map to pave the way towards a completely natural interaction between humans and robots.
This ultimate goal is achieved by defining milestones that promote the use of natural-language interaction.
We underpin our strategy by getting the general public involved in the creation of a large annotated dataset of untrained and unbiased operators, inexistent to the extent of our knowledge.
Here, the competition plays a fundamental role, since \athome
sets the perfect scenario to involve the audience in data production for scientific use.
We believe this data will foster research in both, \nlp and \hri.

The paper is organized as follows.
In \Cref{sec:spr-and-nlp-in-athome}, we briefly introduce the \athome league. In turn, we present a summary of the last nine years of the competition, and give an overview of the strategies used by the teams to solve the tests.
In \Cref{sec:challenges} we summarize the main unresolved challenges in huma-robot interaction.
In \Cref{sec:roadmap} we discuss a roadmap and its milestones to deal with the key problems.
Finally, in \Cref{sec:conclusions} we sum up and draw conclusions.

%
%
%
%
%

\section{Speech Recognition and Natural Language Understanding in \athome}
\label{sec:spr-and-nlp-in-athome}

\athome aims at developing service and assistive robot technology by evaluating a robot's performance with a series of \kword{tests} in an unstandardized and realistic scenario.
Here, new approaches are tested in a competitive setup where robots have to solve a set of common household \kword{tasks} upon request.
Consequently, the competition influences research, even sometimes directing it.
Such is the case of \asr and \nlu, since they are fundamental to \hri.

In a \kword{test}, a task is divided in a sequential set of subgoals or phases, being necessary to fulfill one goal to advance to the next phase~\cite{Iocchi2015}.
This approach keeps difficulty reasonably low for newcomers, while still striving for high top level performance~\cite{Wisspeintner2009}.
Usually, in the beginning, the \kword{operator} gives a spoken command to the robot.
A \kword{command} is typically a short imperative sentence containing all required information to execute a predefined task.
Normally, the operator and the \kword{referee} take account of the robot's processing, whereas the scoring is executed by the \tc[fn], which is also in charge of design tests and review rules.
Detailed descriptions can be found in~\cite{vanderZant2007,Iocchi2015}.

In \athome, \asr and \nlu are closely related.
Spoken commands have always been the preferred way to operate a robot and, since robots need to confirm they have understood the command, both abilities are commonly scored together.

The evolution process of \asr[x] and \nlu[x] in \athome is addressed as follows.
In \Cref{sec:historical-overview}, we summarize the last nine years of competition, focusing on the main challenges the teams had to face.
In \Cref{sec:adopted-strategies}, we present an overview on strategies used to solve the tests.

\subsection{Historical Overview of \hri testing}
\label{sec:historical-overview}
Our review starts in 2009. We have chosen this year because the rulebook of 2009 is the oldest available on the \athome website\footnote{\url{http://www.robocupathome.org}}.

For the competition in \kword{Graz 2009} the guidelines for operating a robot were loose.
Most interactions were hardwired, based on each team's preconceptions of what a natural \hri could be.
In addition, the use of headsets and wireless microphones was common,
and some tests rewarded the use of gestures over speech.

\kword{Istanbul 2010} came with important changes:
\begin{enumerate*}[label=\alph*)]
	\item scoring was made explicit,
	\item score sheets were included in the rulebook,
	and
	\item interaction guidelines were included as part of the tasks in each test.
\end{enumerate*}
For instance, a robot could score for catching the name of a person or room (typically, only the name was given).

But the cornerstone of 2010 was the inclusion of the \gpsr[fn] test.
In this test the operator could command the robot to perform any task from any of other test (including those of former years).
Moreover, robots had to deal with long sentences and incomplete information for the first time; a big step in \nlu and \hri.
Notwithstanding, all interactions in \gpsr followed the patterns used in the other tests.

For \kword{Singapore 2011} most interaction guidelines had been removed and a \kword{command generator}\footnotemark~was developed by the \tc for \gpsr (teams only had access to a limited version).
\footnotetext{Available on-line: \url{http://komeisugiura.jp/software/2010_GeneralPurposeTest.tgz}}

No significant changes in \nlp or spoken \hri were introduced in the next three years.
Nevertheless, the \tc noticed a sustained performance decrease in command retrieval.

To help teams, in \kword{João Pessoa 2014} the \tc introduced a way to bypass speech recognition in order to solve the task: the \kword{\textit{Continue} rule}.
However, only few teams took advantage of it.

For \kword{Hefei 2015} the use of QR codes to bypass \asr was made compulsory.
Moreover, the data recording feature allowed teams to get a scoring bonus and contributing with the league by providing all data acquired by the robot during a test.
In addition, a new command generator was open sourced.
Notwithstanding, even having access to the verbatim output of the \gpsr command generator via the QR Code, many robots remained unresponsive.

In consequence, \kword{Leipzig 2016}, the \tc decided to provide open access to the command generator and the generation grammars about one month before the competition.
This was a crucial decision due to its direct impact on natural \hri, as explained in \Cref{sec:challenges}.
Cloud Computing was another minor but important change.
Prior to this year, the availability of an Internet connection through the arena's wireless network wasn't granted.
However, in combination with the low reliability of the network, discouraged teams from exploring solutions based on cloud services.

By \kword{Nagoya 2017}, the previous two years of tests focused on benchmarking had paid off (See \Cref{tab:speech}).
In addition to an increase in performance in \asr, the relevance of command interpretation and \nlu grew with the inclusion of the \spl.
The out-of-the-box-ready robots allowed teams to focus on high-level problems such as command interpretation, \nlu, and task planning and reasoning.
This year was the first time in which robots had to answer questions about their environment and previously executed tasks.
Moreover, in the Tour-Guide test they also had to attract people outside the competition area, introduce themselves, and answer people's questions without the help of any grammar.

\subsection{Adopted Strategies and Software Solutions for \hri}
\label{sec:adopted-strategies}
Either in face-to-face communication or remotely like by phone, radio, or TV, the hearer decodes the produced sounds of the speaker, trying to match them with the best interpretation given the current context.
Similarly, spoken \hri[x] requires the extraction and analysis of the language elements of the uttered sentence.

\begin{figure}
	\centering
	\vspace{-0.3cm}
	\caption{One of the most common processing chains for spoken commands}
	\label{fig:data-flow}
	\vspace{-3mm}
	\includegraphics[width=0.7\textwidth]{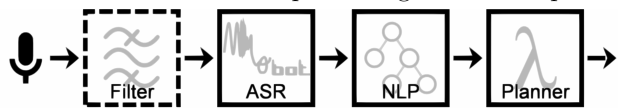}
	\vspace{-0.3cm}
\end{figure}

\begin{wraptable}[15]{i}{0.4\textwidth}
	\centering
	\vspace{-0.7cm}
	\captionsetup{justification=centering}
	\caption{Most used \\ASR software in 2017}
	\label{tab:asr-trends}
	\vspace{-3mm}
	\begin{tabular}{|c|l|}
	\hline
	Usage    & ASR Engine Name        \\
	\hline\hline
	25.81\%  &  CMU [Pocket] Sphinx   \\
	16.13\%  &  Google Speech Api     \\
	16.13\%  &  Microsoft Speech Api  \\
	12.90\%  &  Julius                \\
	 9.68\%  &  Nuance VoCon/DNS      \\
	 6.45\%  &  Rospeex               \\
	 6.45\%  &  Intel RealSense SDK   \\
	 3.23\%  &  Amazon AWS / Alexa    \\
	 3.23\%  &  Kaldi                 \\
	 3.23\%  &  iFlyTek               \\
	 9.68\%  &  Unreported            \\
	\hline
\end{tabular}

\end{wraptable}
Although possible, dealing with audio signals
can be extremely difficult in high levels of abstraction
\cite{Roy2002,Dominey2003}.
Therefore, the most broadly adopted solution consists in using an \asr[fn] engine to get a text-transcript for further processing as depicted in \Cref{fig:data-flow}.
Typically, \kword{cardioid microphones} are used as main audio input for the \asr engine to cope with noise.
Afterwards, the \asr engine output is preprocessed by some sort of [natural] language processor so the task planner can select the most adequate behavior.

Other than the microphone itself, filters are normally absent and the filtering task is delegated to the \asr engine itself~\cite{Doostdar2008}.
Although several noise-reduction filters have been tested in the competitions, we couldn't find any reported successful solution other than \kword{HARK}\footnotemark~\cite{aisltut2017,happymini2017,hibikinoOPL2017}.
However, HARK is used mostly for sound-source localization and separation.
\footnotetext{HARK (\textbf{H}onda Research Institute Japan \textbf{A}udition for \textbf{R}obots with \textbf{K}yoto University) is an open-source robot audition software that includes modules for \asr and sound-source localization and sound separation. Source: \url{https://www.hark.jp/}}

Continuing with the pipeline of \Cref{fig:data-flow},
the most commonly adopted software solutions include Loquendo (now Nuance) ASR~\cite{nimbro2010}, and the Microsoft Speech API~\cite{pumas2013}, being most popular \kword{CMU Sphinx}~\cite{allemaniacs2009,tobi2017} as \Cref{tab:asr-trends} shows.
However, in 2017, due to the limited computing power of the robots in the \spl, most teams used cloud services (mainly the Google speech API~\cite{aisltut2017,jsk2017}) or relied on the built-in \asr system~\cite{uchile2017,utsunleashed2017} which, as the rightmost two columns of \Cref{tab:speech} show, were not as good as other solutions.

\begin{table}
	\centering
	\vspace{-0.7cm}
	\caption{\footnotesize Top-10 scores in ASR-related tests\\
	(final rank in parentheses)}
	\label{tab:speech}
	\begin{tabular}{|c||c|c|c||c|c|}%
	\hline
	\multirow{2}{*}{~Rank~} &
	\multirow{2}{*}{2015} &
	\multirow{2}{*}{2016}               &           \multicolumn{3}{c|}{2017}           \\
	    &  	            &               &      OPL      &     DSPL      &     SSPL      \\
\hline\hline
	 1  &  86.7\% ( 1)  &  86.7\% ( 3)  &  96.7\% ( 1)  &  67.3\% ( 1)  &  71.5\% ( 1)  \\
	 2  &  83.3\% ( 2)  &  83.3\% ( 6)  &  83.3\% ( 3)  &  45.3\% ( 5)  &  52.1\% ( 2)  \\
	 3  &  70.0\% ( 8)  &  72.7\% (13)  &  73.3\% ( 5)  &  37.3\% ( 3)  &  36.4\% ( 4)  \\
	 4  &  70.0\% (10)  &  60.7\% ( 2)  &  70.0\% ( 4)  &  12.0\% ( 4)  &  30.3\% ( 5)  \\
	 5  &  66.7\% ( 6)  &  60.7\% ( 5)  &  66.7\% ( 2)  &   3.3\% ( 7)  &  29.1\% ( 6)  \\
\hline\hline
{\tiny Top5} $\overline{x}$
        &  75.3\%       &  72.8\%       &  78.0\%       &  33.0\%       &  43.9\%       \\
{\tiny Top5} $\sigma$
	    &   9.0\%       &  12.2\%       &  12.2\%       &  25.8\%       &  18.0\%       \\
\hline
\end{tabular}

	\vspace{-0.3cm}
\end{table}

Despite the remarkable advances achieved in \nlp[x] and task planning in recent years, \athome has taken little advantage of it.
Some reasons include
\begin{enumerate*}[label=\alph*)]
	\item the sequential nature of the tests,
	\item the simplicity of the tasks (\hri{}-wise),
	\item the computational power available in the robot,
	\item the lack of awareness due to sensors' limitations,
	and
	\item the need of recognizing only a few words.
\end{enumerate*}
In consequence, most approaches for \nlp relied in \kword{keyword spotting} or pattern matching to trigger the execution of a \kword{state machine}~\cite{Seib2015,golem2013}.
At least in the beginning, this strategy seemed to be faster and more robust than its more advanced counterparts, albeit much simpler.

Despite this, robust A.I. solutions have always been in play.
As of 2013, it was common to find the task planner and the natural language processor fused in the same module, which doesn't seem to be the case anymore.
Common strategies included
\begin{enumerate*}[label=\alph*)]
	\item the use of rules of inference for both, sentence parsing and task planning~\cite{berlinunited2013};
	\item probabilistic parsers~\cite{delft2013},
	\item semantic networks~\cite{pumas2013},
	and
	\item regular expressions~\cite{golem2013}
\end{enumerate*}
to name some.

Although keyword spotting, \kword{pattern matching}, and state machines are currently used solutions (specially in simple tests), we spotted more powerful and avant-grade approaches.
Whilst in 2017 only 39\% of the teams didn't mentioned neither \nlp nor \nlu approaches in their \tdp, in all other reports the task planner and the \nlp are separated.
Among the approaches for processing language, we found
\begin{enumerate*}[label=\alph*)]
	\item Probabilistic Semantic Parsers~\cite{utaustinvilla2017};
	\item Multimodal Residual Deep Neural Networks~\cite{wrighteagle2017},
	\item ontology-based parsers over inference engines~\cite{pumas2017},
	and
	\item probabilistic parsers for syntax-tree extraction along with lambda calculus for semantic parsing~\cite{aupair2017}.
\end{enumerate*}
The \kword{Stanford Parser}~\cite{stanford2011corenlponline} is the most broadly adopted solution for POS-tagging and syntactic tree extraction, and \kword{LU4R}~\cite{Bastianelli2016}, a Spoken Language Understanding Chain for HRI developed in La Sapienza~\cite{spqrel2017} by participants of \athome which is used by several teams.

These newer solutions have increased the robots' performance in the latter versions of \gpsr when some ambiguity was added.
Nevertheless, the league does not acquire these newer approaches.
Most robots are still using grammar-based \asr engines, which limit the input of the \nlp unit.
Moreover, so far no test requires the resolution of ellipsis or anaphora.
Finally, another major inconvenience, is the use of command generators instead of natural interactions.
How to develop \hri tests which focus on these problems is discussed in \Cref{sec:challenges}.

%
%
%
%
%

\section{Challenges}
\label{sec:challenges}

As stated in previous sections, several setbacks have been holding back the league's advances in natural language dialogs with the robot.
Here we exemplify challenges and come up with guidelines for future research when addressing tougher scenarios.

\subsection{Noise}
One of the most problematic aspects (and to which most attention has been paid) is the noise in the competition environment.
Having a separated hall as in 2012 didn't help much.
The \kword{ambient noise} produced by over two hundred people greatly exceeds the noise levels of an average apartment.

In the past, several solutions have been proposed to this problem.
Having a separate hall, arenas with tall walls, and transform the arena in a sound-proof closed area with glass walls are recurrent examples.
However, service robots will also operate in noisy environments such as airports and shopping malls.
Thus, we think it is best to deal with this issue in an early stage.
For this reason, noise is addressed in the roadmap presented in \Cref{sec:roadmap}.

\subsection{Operators}
An aspect that has characterized \athome is that the league has always been robot-friendly.
People volunteering as operators in \athome are often patient and prone to follow the robot's instructions, unconsciously trying to help the robot to succeed (e.g. repeat a given command louder when the robot seems unresponsive).
In addition, almost all the operators are specialists, or at least familiar with service robotics.
In other words, all testing has been performed by (unconsciously) \kword{biased operators}, making it easier for the robots to accomplish their tasks.
However, trying to give a positive impression to the audience, has the disadvantage of voiding the purpose of \athome of providing real-case scenarios for testing.

Another important closely related aspect addresses the demographics of the league's participants.
An international community will offer \kword{diverse accents} to the robots and styles (which may not be correct).
But this diversity comes with a price.
With few exceptions, operators having English as second language lack the richness and verbal fluency of a native speaker.
Same hold for gender and age.

Therefore bias, variety of speech and diversity of accents, lexicon, and styles, are also addressed in the roadmap of \Cref{sec:roadmap}.

\subsection{Generators}
Although the rulebooks never provided interaction templates, suggested guidelines and the sentences produced by the command generators codify biased \hri in a similar way.
Besides, officializing the release of the command generator and its \kword{grammars} had the pernicious effect of replacing natural language with a simplistic one, i.e.~artificializing interactions.
Moreover, the person in charge of the generator (one of the authors of this paper), although proficient, is not a native speaker of the official competition language (US-English).
In summary, despite stimulating immediate positive results and supplying the teams with a powerful tool, the careless use of generators might delve into a ballast in the long term.
For this reason, the use of command generators is revised in the roadmap of \Cref{sec:roadmap}.

In order to overcome the problems discussed in this section, the proposed solution should force robots to deal with a vast diversity of unbiased operators speaking freely and under reasonable conditions of noise,
while keeping tests fair and scientifically meaningful.
This ultimate goal is analyzed and split into small steps which are presented in \Cref{sec:roadmap}.
%
%
%
%
%
\section{Solution strategy and roadmap}
\label{sec:roadmap}

Ideally, the robots will lead a natural-language dialogue in real-environment conditions.
This is the ultimate goal in service robotics regarding \nlu.
Considering the challenges presented in \Cref{sec:challenges} we propose a series of specific tests and changes to the existing ones.

In the previous section we identified the biased operators in combination with restricted GPSR as a main obstacle for the stagnation in elaborate HRI.
In order to set up a rich corpus of human-robot dialogs, we opt for recruiting untrained English native speakers as \kword{untrained operators}, e.g.~selected from the audience.
The \kword{unbiased operators} must have little to no previous experience in \athome (or preferably in robotics), and be allowed to interact freely with the robot.
With these changes, we are solving biasing, variety and diversity of speech, and tackling the artificiality introduced with the command generators.

Needless to say, the proposed changes  towards free natural-language dialogs with the robot have to be applied gradually.
Therefore, in this section we present a roadmap of three phases that takes as axis the \gpsr[xn] test before changes are propagated and adapted for other tests.
Each phase sets a constraint that will rule over the upcoming featured milestones.
Milestones present \kword{small increments} in the difficulty of the \nlu task, presumably achievable with the data collected in former years.
Once solved in \gpsr, the constraints of a milestone would become standard practices in other tests.
In addition, milestones are planned to respect the two-years-limit established by the founders of the league~\cite{Iocchi2015}.
Furthermore, the proposed roadmap has the advantage of being adaptable.
If sufficient performance hasn't been achieved, the milestones can be shifted forward in time.

In order not to overtax the teams, we propose the following 3-step pipeline to implement the changes by small increments towards free natural-language dialogs with the robots.
First, teams test and benchmark with recordings addressing the features of the milestones newly defined; then, those features are tested in \gpsr; and finally, propagated to all other pertinent tests.

Hence, the strategy considers also the inclusion of a \textit{\nlu[x] and Action Planning} benchmark in Stage~I with a small contribution to the overall score.
In this benchmark, a team receives a set of \kword{sound files}, each one containing a task-execution request from an unbiased operator of the same kind that would be given in \gpsr.
The A.I. of the robot needs to transcribe and analyze these recordings, extracting either a goal, a plan with a set of actions to carry out, or a set of questions or statements to continue the interaction.
The score should consider not only the analytical quality of producing transcripts (where applicable penalizing hard-wired constructions in favor of generalized \nlp-rules), but also how far the robot went in its planning, and if it was following the right direction.
In this way, milestones could be tested one or two years before being implemented in \gpsr and propagated to other tests, giving time to teams to prepare.
Furthermore, by giving the same recordings to all teams, \kword{fairness} and replicability are addressed,
while noise can be tackled by superposing ambient \kword{noise} to the recordings of the operators.
It will be up to the \tc to decide on the nature and intensity of the noise regarding the league's advances in the subject.

In addition, to support the league and foster research,
the proposed strategy exploits the \kword{Data Recording}
feature, incorporated in 2015 to the competition. This is deemed as fundamental for the success of our proposal.
We think all interactions between the operator and the robot should be recorded and distributed under \kword{Open Access} license as soon as a transcription is available.
This includes all benchmarking sets from previous years.
In this way, the league supports both, experienced and new competitors.
Beyond this obvious assistance the entire scientific community gains relevance.
In early stages, teams providing recordings of all speech-based interactions during a test might receive a bonus proportional to the score obtained, as well as an additional bonus or a certificate (as decided by the \tc) for annotating the provided data.
Later on, such policy should be made compulsory for the roadmap to work with full efficiency, automating the collection process if possible.

Before summing up by means of presenting the roadmap, we have to state some minor clarifications:
\begin{enumerate*}[label=\arabic*)]
	\item In all phases, operators with no background in robotics are selected from the audience by the \tc.
	\item Unless the test specifies otherwise, chosen operators shall be fluent English speakers.
	\item Commands, goals, and tasks are generated before the test. A referee must explain all pertinent information to the operators before spoken interactions are recorded and noise overlapped.
	\item The \tc randomly assigns to each robot a set of generated tasks from the pool. The robot will listen to the recording, but the operator must be present for further (unexpected) interactions.
\end{enumerate*}

~\\\noindent\textbf{Roadmap phase 1: From commands to goal formulation}

The referee gives a specific goal to the operator along with all pertinent information regarding the desired task.
For this purpose, a random command generator can be used.
When required, the operator can practice with the referee. Referees assist operators in unexpected situations.

The intensity of overlapped noise should increase gradually, starting from relatively quiet environments (country house, city apartment, office, etc) towards moderately noisy ones (busy office, or a restaurant with background music).

~\\\noindent\textbf{Yearly progress}
\begin{enumerate}[nosep, leftmargin=1cm, rightmargin=1cm]
\footnotesize
	\item[2019] The operator reads the generated command.\\
	\textbf{Note:} This vanilla milestone introduces no change to allow the benchmarking of the next milestone in the pipeline.
	\item[2020] The operator tries to memorize the command, repeating it to the robot afterwards. Slight variations are expected.
	\item[2022] The referee explains the task to the operator, who has to command the robot using their own words (e.g. rephrasing).
	\item[2024] The operator requires the robot to accomplish a task, although not necessarily in an imperative way, as if suggesting.
	\item[2028] The operator tries to explain the goal to the robot, not specifying what to do, but the expected final result.
\end{enumerate}

While no big changes are expected during the first four years in \nlp, we foresee the inclusion of more robust filters and the use of less constrained grammars.
However, by the third milestone (2022), operators might unintentionally neglect information, so robots will need to reconcile information as it arrives, making \kword{questions} as needed.
Furthermore, people normally make a very efficient use of language, so references are very common.
Therefore, we expect the exploration of \kword{reference resolution} such as ellipsis and anaphora by this year.
Finally, the latter milestones will keep the pace, slightly moving the focus to task planning while addressing new types of sentences.

~\\\noindent\textbf{Roadmap phase 2: Towards dialog-based interaction}

The referee gives to the operator a set of examples of what robots can do in a given domain. The operator must propose a similar task for the robot which has to be approved by the referee. The robot may not know how to accomplish the task, in which case the procedure is explained by the operator.

The intensity of overlapped noise ranges from medium to high, with sudden bursts of loud recorded human voices like in shopping malls, grocery stores, and airports are good examples.

~\\\noindent\textbf{Yearly progress}
\begin{enumerate}[nosep, leftmargin=1cm, rightmargin=1cm]
\footnotesize
	\item[2032] After receiving a set of examples, the operator elicits a similar behavior to the robot.
	The procedure can be explained in detail step by step.
	\item[2036] After receiving a topic and a set of examples, the operator requests something of the same difficulty.
	Sub-goals are detailed to the robot, but not the individual commands.
	The operator corrects the robot's plan.
\end{enumerate}

In these phases, we continuously add new elements to \nlu, integrating it deeper with \kword{planning}, while, at the same time, we are collecting new applications from users (the potential market).
Plan-learning using natural language requires the integration of \kword{cardinality}, as well as spatial and temporal relationships
(e.g. the operator may request \textit{make me a sandwich}, explaining later on the steps like \textit{grab a slice of bread, spread mayo\dots}).

In contrast to only presenting an initiated plan to the operator, the tasks stimulate that the operator and the robot have to enter in a \kword{dialog} towards an unconstrained natural interaction.

~\\\noindent\textbf{Roadmap Goal: Reaching unconstrained Interaction}

In the last phase, the operators only get explained the State-of-the-Art, i.e. the limits of what the robots can achieve.
They can request anything crossing their minds within those limits the way they want. The robot may not know how to accomplish the task, in which case the procedure is freely explained by the operator in a dialog.

~\\\noindent\textbf{Yearly progress}
\begin{enumerate}[nosep, leftmargin=1cm, rightmargin=1cm]
\footnotesize
	\item[2040] The operator explains to the robot how to perform an entirely new task and what the results should be.
	\item[2044] The operator requires anything from the robot within the state-of-the-art. All planning is left to the robot.
\end{enumerate}
~\\
We believe the presented roadmap helps the \athome community to push the boundaries of research in \hri.

The first steps will force teams to look for alternatives to grammar-based \asr engines, or at least a less restricted ones.
Besides, the continuous analysis of audio signals to separate the operators' voice from noise is also addressed, although dealing with noise can be left as an option for daring teams.

Regarding \nlu, the first steps will take current approaches to the limit.
Unbiased operators will gradually expose robots to the richness of \kword{free speech}, while the pipeline will grant time to prepare.

Later on, moving from imperative sentences (e.g. \textit{clean the bedroom}) to goal-driven sentences of any kind (e.g. \textit{I need the room clean for tonight}, \textit{is the breakfast ready?}) will not only have the league working with semantics and pragmatics, but also might foster research in action planning.

Especially, in the second phase, more complex elements of language analysis are incorporated, paving the road towards a dialog-based interaction.
On that basis, we stipulate conversational robots in \athome.
This claim is substantiated in the last phase by widening the domain and removing all constrains from the operators, in order to deal with real-world situations.

Nevertheless the roadmap provides the necessary flexibility for the worst-case scenario where direct instructions can be given to the robots.

%
%
%
%
%

\section{Conclusions}
\label{sec:conclusions}

In this paper we provided a historical overview of testing natural-language interactions with robots over the last nine years of \athome.
We outlined the state-of-the-art \asr[x] and \nlp[x].
We quantified recent strategies and software solutions adopted by teams to overcome the trials set in the competition.

In these observations we identified a set of challenges that haven't been tackled.
Critical components that hamper free dialogs with the robot are posed by the command generator along with the subconscious unintentional bias of the operators.
We inspected how individual test features prevent the league from dealing with free natural language.
Based on this study, we propose a strategy and a roadmap that formulate key features to implement the in the individual \nlp components of the robot (see~\Cref{fig:data-flow}) to conquer those challenges.
Many test details strongly depend on the advancement of \asr engines, neglects non-verbal communication, so there should be an entangled roadmap for these features.

We hope this roadmap will serve as initiative to promote long-term planning in \athome.
In particular our contribution elaborates on stepstones to elicit thorough \nlp to empower robots with strong natural-language skills.

\bibliographystyle{bst/splncs04}
\scriptsize
\bibliography{bib/paper,bib/tdps2017}

\begin{thebibliography}{10}
\providecommand{\url}[1]{\texttt{#1}}
\providecommand{\urlprefix}{URL }
\providecommand{\doi}[1]{https://doi.org/#1}

\bibitem{Bastianelli2016}
Bastianelli, E., Croce, D., Vanzo, A., Basili, R., Nardi, D.: A discriminative
  approach to grounded spoken language understanding in interactive robotics.
  In: IJCAI. pp. 2747--2753 (2016)

\bibitem{happymini2017}
Demura, K., Demura, K., Nagashima, K., Enomoto, K., Yamakawa, T., Iwasaki, R.,
  Mashimo, S.: Happy mini 2017 team description paper. RoboCup @Home 2017 Team
  Description Papers  (2017)

\bibitem{Dominey2003}
Dominey, P.F.: Learning grammatical constructions from narrated video events
  for human--robot interaction. In: Proceedings IEEE humanoid robotics
  conference, Karlsruhe, Germany (2003)

\bibitem{Doostdar2008}
Doostdar, M., Schiffer, S., Lakemeyer, G.: A robust speech recognition system
  for service-robotics applications. In: Iocchi, L., Matsubara, H.,
  Weitzenfeld, A., Zhou, C. (eds.) RoboCup 2008: Robot Soccer World Cup XII.
  pp. 1--12. Springer Berlin Heidelberg, Berlin, Heidelberg (2009)

\bibitem{delft2013}
Gaisser, F., Aswin~Chandarr, A., Rudinac, M., Bruinink, M., Pons, S., Rueda,
  M.B., Lung, G.L., Wisse, M., Jonker, P.: Delft robotics robocup@home 2013
  team description paper. Proceedings RoboCup competition  (2013)

\bibitem{utaustinvilla2017}
Hart, J.W., Stone, P., Thomaz, A., Niekum, S.: Ut austin villa robocup@home
  domestic standard platform league team description paper. RoboCup @Home 2017
  Team Description Papers  (2017)

\bibitem{hibikinoOPL2017}
Hori, S., Ishida, Y., Kiyama, Y., Tanaka, Y., Kuroda, Y., Hisano, M., Imamura,
  Y., Himaki, T., Yoshimoto, Y., Aratani, Y., Hashimoto, K., Iwamoto, G.,
  Morie, T., Tamukoh, H.: Hibikino-musashi@home 2017 team description paper.
  RoboCup @Home 2017 Team Description Papers  (2017)

\bibitem{Iocchi2015}
Iocchi, L., Holz, D., Ruiz-del Solar, J., Sugiura, K., Van Der~Zant, T.:
  Robocup@home: Analysis and results of evolving competitions for domestic and
  service robots. Artificial Intelligence  \textbf{229},  258--281 (2015)

\bibitem{aupair2017}
Lee, B.J., Choi, J.Y., Lee, C.Y., Park, K.W., Choi, S., Baek, C., Zhang, B.T.:
  2017 aupair team description paper. RoboCup @Home 2017 Team Description
  Papers  (2017)

\bibitem{wrighteagle2017}
Liu, J., Zhang, Z., Tang, B., Chen, X.: Wrighteagle@home 2017 team description
  paper. RoboCup @Home 2017 Team Description Papers  (2017)

\bibitem{berlinunited2013}
Llarena, A., Boldt, J.F., Steinke, N.S., Engelmeyer, H., Rojas, R.:
  Berlinunited@ home 2013 team description paper. Proceedings RoboCup
  competition  (2013)

\bibitem{uchile2017}
Mart{\'i}nez, L., Mu{\~n}oz, R., Olave, G., Pais, G., Hernan, G., Gomez, D.,
  Garrido, L., Campanini, D., Orellana, P., Loncomilla, P., Ruiz-del Solar, J.:
  Uchile homebreakers 2017 team description paper. RoboCup @Home 2017 Team
  Description Papers  (2017)

\bibitem{spqrel2017}
M.T.~Lázaro, Iocchi, L., Nardi, D., Hanheide, M., Fentanes, J.P.: Spqrel 2017
  team description paper. RoboCup @Home 2017 Team Description Papers  (2017)

\bibitem{aisltut2017}
Oishi, S., Miura, J., Koide, K., Demura, M., Kohari, Y., Une, S.,
  Villamar-Gomez, L., Kato, T., Kojima, M., Morohashi, K.: Aisl-tut @home
  league 2017 team description paper. RoboCup @Home 2017 Team Description
  Papers  (2017)

\bibitem{golem2013}
Pineda, L., Meza, I., Fuentes, G., Rasc{\'o}n, C., Pe{\~n}a, M., Ortega, H.,
  Reyes-Castillo, M., Salinas, L., Ortega, J., Rodr{\'\i}guez-Garc{\'\i}a, A.,
  et~al.: The golem team, robocup@home 2013 (2013)

\bibitem{Roy2002}
Roy, D.K., Pentland, A.P.: Learning words from sights and sounds: A
  computational model. Cognitive science  \textbf{26}(1),  113--146 (2002)

\bibitem{pumas2013}
Savage, J., Negrete, M., Matamoros, M., Cruz, J., Contreras, L., Pacheco, A.,
  Figueroa, I., M{\'a}rquez, J.: Pumas @home 2013 team description paper (2013)

\bibitem{pumas2017}
Savage, J., Negrete, M., Cruz, J., Marquez, J., Martell, R., Cruz, J., Vazquez,
  E., Pano, M., Cruz, J., Silva, E., Estrada, H., Arce, H., Matamoros, M.,
  Garzon, A., Fuentes, O.: Pumas@home 2017 team description paper. RoboCup
  @Home 2017 Team Description Papers  (2017)

\bibitem{allemaniacs2009}
Schiffer, S., Niem{\"u}ller, T., Doostdar, M., Lakemeyer, G.: Allemaniacs @home
  2009 team description. Proceedings CD RoboCup  (2009)

\bibitem{Seib2015}
Seib, V., Manthe, S., Memmesheimer, R., Polster, F., Paulus, D.: Team
  homer@unikoblenz—approaches and contributions to the robocup@home
  competition. In: Robot Soccer World Cup. pp. 83--94. Springer (2015)

\bibitem{stanford2011corenlponline}
Stanford: Corenlp (2011), \url{http://nlp.stanford.edu:8080/corenlp/}

\bibitem{nimbro2010}
St{\"u}ckler, J., Dr{\"o}schel, D., Gr{\"a}ve, K., Holz, D., Schreiber, M.,
  Behnke, S.: Nimbro @home 2010 team description (2010)

\bibitem{tobi2017}
Wachsmuth, S., Lier, F., Meyer~zu Borgsen, S., Kummert, J., Lach, L., Sixt, D.:
  Tobi - team of bielefeld a human-robot interaction system for robocup@home
  2017. RoboCup @Home 2017 Team Description Papers  (2017)

\bibitem{utsunleashed2017}
Williams, M.A., Pfeiffer, S., Vitale, J., Tonkin, M., Ojha, S., Billingsley,
  R., Alam, S., Kang, L., Gudi, S.L.K.C., Clark, J., Wang, X., Johnston, B.:
  Uts unleashed! for robocup 2017 @home spl. RoboCup @Home 2017 Team
  Description Papers  (2017)

\bibitem{Wisspeintner2009}
Wisspeintner, T., Van Der~Zant, T., Iocchi, L., Schiffer, S.: Robocup@home:
  Scientific competition and benchmarking for domestic service robots.
  Interaction Studies  \textbf{10}(3),  392--426 (2009)

\bibitem{jsk2017}
Yaguchi, H., Tran, B., Takeda, M., Kochigami, K., Li, Z., Sasabuchi, K.,
  Furuta, Y., Nagahama, K., Okada, K., Inaba, M.: Jsk@home: Team description
  paper for robocup@home 2017. RoboCup @Home 2017 Team Description Papers
  (2017)

\bibitem{vanderZant2007}
van~der Zant, T., Wisspeintner, T.: Robocup@home: Creating and benchmarking
  tomorrows service robot applications. In: Robotic Soccer. InTech (2007)

\end{thebibliography}

\end{document}